\def\year{2018}\relax
\documentclass[letterpaper]{article} 
\usepackage{aaai18}  
\usepackage{times}  
\usepackage{helvet}  
\usepackage{courier}  
\usepackage{url}  
\usepackage{graphicx}  
\usepackage{latexsym}
\usepackage{amsmath}
\usepackage{multirow}
\usepackage{bm}
\begin{document}
%
\title{Emotional Chatting Machine: Emotional Conversation Generation with Internal and External Memory}
\author{Hao Zhou$^\dagger$, \quad Minlie Huang$^\dagger$\thanks{Corresponding author: Minlie Huang, aihuang@tsinghua.edu .cn}, \quad Tianyang Zhang$^\dagger$, \quad Xiaoyan Zhu$^\dagger$, \quad Bing Liu$^\ddagger$ \\
  $^\dagger$State Key Laboratory of Intelligent Technology and Systems, \\
  National Laboratory for Information Science and Technology, \\
  Dept. of Computer Science and Technology, Tsinghua University, Beijing 100084, PR China \\
  $^\ddagger$Dept. of Computer Science, University of Illinois at Chicago, Chicago, Illinois, USA \\
  {\tt tuxchow@gmail.com} ,\quad {\tt aihuang@tsinghua.edu.cn} , \\ \quad {\tt keavilzhangzty@gmail.com}, \quad {\tt zxy-dcs@tsinghua.edu.cn}, \quad {\tt liub@cs.uic.edu}}
\maketitle
\begin{abstract}
Perception and expression of emotion are key factors to the success of dialogue systems or conversational agents. However, this problem has not been studied in large-scale conversation generation so far. In this paper, we propose Emotional Chatting Machine (ECM) that can generate appropriate responses not only in content (relevant and grammatical) but also in emotion (emotionally consistent).
To the best of our knowledge, this is the first work that addresses the emotion factor in large-scale conversation generation. 
ECM addresses the factor using three new mechanisms that respectively (1) models the high-level abstraction of emotion expressions by embedding emotion categories, (2) captures the change of implicit internal emotion states, and (3) uses explicit emotion expressions with an external emotion vocabulary.
Experiments show that the proposed model can generate responses appropriate not only in content but also in emotion.

\end{abstract}
\section{Introduction} 
\label{sec:introduction}

As a vital part of human intelligence, emotional intelligence is defined as the ability to perceive, integrate, understand, and regulate emotions~\cite{salovey1997emotional}. It has been a long-term goal of artificial intelligence to enable a machine to understand affect and emotion~\cite{picard1997affective}. To create a chatbot capable of communicating with a user at the human level, it is necessary to equip the machine with the ability of perceiving and expressing emotions.

Existing studies show that addressing affect and emotion in dialogue systems or conversational agents can enhance 
user satisfaction \cite{prendinger2005using}, 

and lead to fewer breakdowns in dialogues~\cite{martinovski2003breakdown}. Some initial work has also been done on adjusting dialogue behaviors to suit users' emotional states 
\cite{polzin2000emotion}, and on generating responses to users' utterances at both the content- and affect-related levels \cite{skowron2010affect}.

However, these studies, mostly inspired by psychology findings, are either rule-based or limited to small-scale data. Recently, neural models trained on large-scale data have advanced open-domain conversation generation significantly~\cite{Ritter2011Data,Vinyals2015A,Shang2015Neural,Serban2016Building}. 
Most of these models aim to improve the {\bf content quality} of conversation generation~\cite{gu2016CopySeq2Seq,li2015diversityMMI,xing2016topicAware,mou2016sequenceBackwardForward,LiGBSGD16}. 
To the best of our knowledge, the \textbf{emotion factor} has not been addressed in existing neural models for conversation generation. Table \ref{tab:example} shows some examples with/without considering emotions. We can see that our ECM model is emotionally involved and more empathetic. 

\begin{table} [!htp]
    \centering
    \begin{tabular}{|l|}
        \hline
        User: Worst day ever. I arrived late because of the traffic. \\ 
        \hline 
        Basic Seq2Seq: You were late. \\ 
        ECM ({\it Like}): I am always here to support you. \\ 
        ECM ({\it Happy}): Keep smiling! Things will get better. \\
        ECM ({\it Sad}): It's depressing. \\ 
        ECM ({\it Disgust}): Sometimes life just sucks. \\ 
        ECM ({\it Angry}): The traffic is too bad! \\ 
         
        \hline 
    \end{tabular} 
    \caption{Conversations with/without considering emotion.}
    \label{tab:example}
\end{table}

There are several challenges in addressing the emotion factor in large-scale conversation generation. {\bf First}, high-quality emotion-labeled data are difficult to obtain in a large-scale corpus, as emotion annotation is a fairly subjective task and emotion classification is also challenging.  
{\bf Second}, it is difficult to consider emotions in a natural and coherent way because we need to balance grammaticality and expressions of emotions, as argued in \cite{GhoshCLMS17}. {\bf Last}, { simply embedding emotion information in existing neural models, as shown in our experiments, cannot produce desirable emotional responses but just hard-to-perceive general expressions (which contain only common words that are quite implicit or ambiguous about emotions, and amount to 73.7\% of all emotional responses in our dataset).
}

In this paper, we address the problem of generating emotional responses in open-domain conversational systems and propose an emotional chatting machine (ECM for short). To obtain large-scale emotion-labeled data for ECM, we train a neural classifier on a manually annotated corpus. The classifier is used to annotate large-scale conversation data automatically for the training of ECM. To express emotion naturally and coherently in a sentence, we design a sequence-to-sequence generation model equipped with new mechanisms for emotion expression generation, namely, 
emotion category embedding for capturing high-level abstraction of emotion expressions, an internal emotion state for balancing grammaticality and emotion dynamically, and an external emotion memory to help generate more explicit and unambiguous emotional expressions.

In summary, this paper makes the following contributions:
\begin{itemize}
    \item It proposes to address the emotion factor in large-scale conversation generation. To the best of our knowledge, this is the first work on the topic.
    
    \item It proposes an end-to-end framework (called ECM) to incorporate the emotion influence in large-scale conversation generation. 
    It has three novel mechanisms: emotion category embedding, an internal emotion memory, and an external memory.
       
    \item It shows that ECM can generate responses with higher content and emotion scores than the traditional seq2seq model. We believe that future work such as the empathetic computer agent and the emotion interaction model can be carried out based on ECM.

\end{itemize}

\section{Related Work} 
\label{sec:related_work}

In human-machine interactions, 
the ability to detect signs of human emotions and to properly react to them can enrich communication. 
For example, display of empathetic emotional expressions enhanced users' performance \cite{partala2004effects}, and led to an increase in user satisfaction \cite{prendinger2005using}. 
Experiments in \cite{prendinger2005empathic} showed that an empathetic computer agent can contribute to a more positive perception of the interaction. 
In \cite{martinovski2003breakdown}, the authors showed that many breakdowns could be avoided if the machine was able to recognize the emotional state of the user and responded to it sensitively. 
The work in \cite{polzin2000emotion} presented how dialogue behaviors can be adjusted to users' emotional states. Skowron \shortcite{skowron2010affect} proposed  conversational systems, called affect listeners, that can respond to users' utterances both at the content- and affect-related level.

These works, mainly inspired by psychological findings, are either rule-based, or limited to small data, making them difficult to apply to large-scale conversation generation.
Recently, sequence-to-sequence generation models~\cite{sutskever2014sequence,bahdanau2014neural} have been successfully applied to large-scale conversation generation~\cite{Vinyals2015A}, including neural responding machine \cite{Shang2015Neural}, hierarchical recurrent models \cite{serban2015hierarchical}, and many others.
These models focus on improving the content quality of the generated responses, including diversity promotion \cite{li2015diversityMMI}, considering additional information  \cite{xing2016topicAware,mou2016sequenceBackwardForward,LiGBSGD16,herzig2017neural}, and handing unknown words~\cite{gu2016CopySeq2Seq}.

However, no work has addressed the emotion factor in large-scale conversation generation.
There are several studies that generate text from controllable variables. 
\cite{hu2017toward} proposed a generative model which can generate sentences conditioned on certain attributes of the language such as sentiment and tenses. Affect Language Model was proposed in \cite{GhoshCLMS17} to generate text conditioned on context words and affect categories.
\cite{cagan2017data} incorporated the grammar information to generate comments for a document using sentiment and topics.
Our work is different in two main aspects: 1) prior studies are heavily dependent on linguistic tools or customized parameters in text generation, while our model is fully data-driven without any manual adjustment; 2) prior studies are unable to model multiple emotion interactions between the input post and the response, instead, the generated text simply continues the emotion of the leading context.

\begin{figure*}[!htp]
  \centering
  \includegraphics[width=1.0\linewidth]{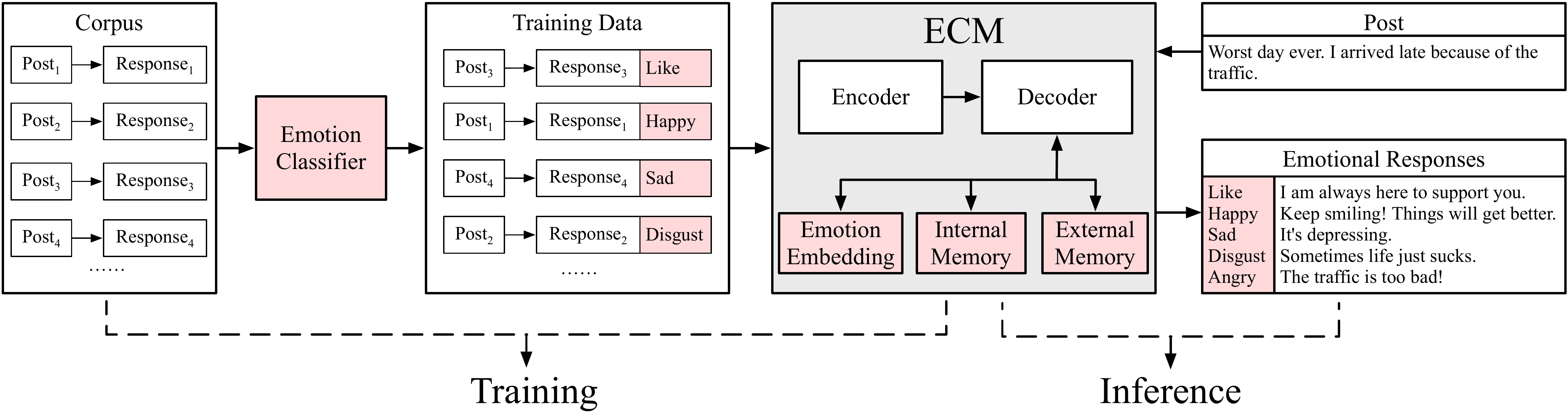}
  \caption{Overview of ECM (the grey unit). The pink units are used to model emotion factors in the framework.}
  \label{fig:1}
\end{figure*}

\section{Emotional Chatting Machine} 

\subsection{Background: Encoder-decoder Framework}

Our model is based on the encoder-decoder framework of the general sequence-to-sequence (seq2seq for short) model \cite{sutskever2014sequence}. It is implemented with gated recurrent units (GRU) \cite{cho2014learning,chung2014empirical}.
The encoder converts the post sequence $\bm{X}= (x_1,x_2,\cdots, x_{n})$ to hidden representations $\bm{h} = (\bm{h}_1,\bm{h}_2,\cdots, \bm{h}_n)$, which is defined as:
\begin{equation}
\bm{h}_t = \mathbf{GRU}(\bm{h}_{t-1},x_t) .
\end{equation}

The decoder takes as input a context vector $\bm{c}_t$ and the embedding of a previously decoded word $\bm{e}(y_{t-1})$ to update its state $\bm{s}_t$ using another GRU:
\begin{eqnarray}
\bm{s}_t  & = & \mathbf{GRU}(\bm{s}_{t-1},[\bm{c}_t ; \bm{e}(y_{t-1})]) ,
\label{eq:decoder}
\end{eqnarray}
where $[\bm{c}_t; \bm{e}(y_{t-1})]$ is the concatenation of the two vectors, serving as the input to the GRU cell. 
The context vector $\bm{c}_t$ is designed to dynamically attend on key information of the input post during decoding~\cite{bahdanau2014neural}. 
Once the state vector $\bm{s}_t$ is obtained, the decoder generates a token by sampling from the output probability distribution $\bm{o}_t$ computed from the decoder's state $\bm{s}_t$  as follows: 
\begin{eqnarray}
y_t \sim \bm{o}_t & = & P(y_t \mid y_1, y_2,\cdots, y_{t-1}, \bm{c}_t) ,\\
\label{eq:soft}
& = & {\rm softmax}(\mathbf{W_o} \bm{s}_t) .
\end{eqnarray}

\subsection{Task Definition and Overview}

Our problem is formulated as follows: Given a post $\bm{X} = (x_1,x_2,\cdots, x_{n})$ and an 
emotion category $e$ of the response to be generated (explained below), the goal is to generate a response $\bm{Y} = (y_1,y_2,\cdots, y_{m})$ that is coherent with the emotion category $e$. Essentially, 
the model estimates the probability:
$P(\bm{Y}|\bm{X},e)=\prod_{t=1}^{m}P(y_t|y_{<t},\bm{X},e)$.
The emotion categories are \{{\it Angry, Disgust, Happy, Like, Sad, Other}\}, adopted from a Chinese emotion classification challenge task.\footnote{The taxonomy comes from http://tcci.ccf.org.cn/confere

-nce/2014/dldoc/evatask1.pdf} 

In our problem statement, we assume that the emotion category of the to-be-generated response is given, because emotions are highly subjective. Given a post, there may be multiple emotion categories that are suitable for its response, depending on the attitude of the respondent.
For example, for a sad story, someone may respond with sympathy (as a friend), someone may feel angry (as an irritable stranger), yet someone else may be happy (as an enemy). Flexible emotion interactions between a post and a response are an important difference from the previous studies \cite{hu2017toward,GhoshCLMS17,cagan2017data}, which use the same emotion or sentiment for response as that in the input post.

Thus, due to this subjectivity of emotional responses, we choose to focus on solving the {\bf core problem}: generating an emotional response given a post and an emotion category of the response. Our model thus works regardless the response emotion category. Note that there can be multiple ways to enable a chatbot to choose an emotion category for response. One way is to give the chatbot a personality and some background knowledge. Another way is to use the training data to find the most frequent response emotion category for the emotion in the given post and use that as the response emotion. This method is reasonable as it reflects the general emotion of the people. We leave this study to our future work.

Building upon the generation framework discussed in the previous section, 
we propose the Emotional Chatting Machine (ECM) to generate emotion expressions using three mechanisms:
{\bf First}, since the emotion category is a high-level abstraction of an emotion expression, ECM embeds the emotion category and feeds the emotion category embedding to the decoder. {\bf Second}, we assume that during decoding, there is an internal emotion state, and in order to capture the implicit change of the state and to balance the weights between the grammar state and the emotion state dynamically, ECM adopts an internal memory module. {\bf Third}, an explicit expression of an emotion is modeled through an explicit selection of a 
generic (non-emotion) or emotion word by an external memory module.

An overview of ECM is given in Figure \ref{fig:1}. 
In the training process, the corpus of post-response pairs is fed to an emotion classifier to generate the emotion label of each response, and then ECM is trained on the data of triples: posts, responses and emotion labels of responses. In the inference process, a post is fed to ECM to generate emotional responses conditioned on different emotion categories.

\subsection{Emotion Category Embedding}
Since an emotion category (for instance, {\it Angry, Disgust, Happy}) provides a high-level abstraction of an emotion expression,
the most intuitive approach to modeling emotion in response generation is to take as additional input the emotion category of a response to be generated. 
Each emotion category is represented by a real-valued, low dimensional vector. For each emotion category $e$, we randomly initialize the vector of an emotion category $\bm{v}_e$, and then learn the vectors of the emotion category through training. The emotion category embedding $\bm{v}_e$, along with word embedding $\bm{e}(y_{t-1})$, and the context vector $\bm{c}_t$, are fed into the decoder to update the decoder's state $\bm{s}_t$:
\begin{equation}
\bm{s_t} = \mathbf{GRU}(\bm{s}_{t-1}, [\bm{c}_t ; \bm{e}(y_{t-1}) ; \bm{v}_e]) .
\end{equation}
Based on $\bm{s}_t$, the decoding probability distribution can be computed accordingly by Eq. \ref{eq:soft} to generate the next token $y_t$.

\subsection{Internal Memory}
\label{sec:internal}

The method presented in the preceding section is rather {\bf static}: the emotion category embedding will not change during the generation process which may sacrifice grammatical correctness of sentences as argued in \cite{GhoshCLMS17}. Inspired by the psychological findings that emotional responses are relatively short lived and involve changes \cite{gross1998emerging,hochschild1979emotion}, and the dynamic emotion situation in emotional responses \cite{alam2017annotating}, we design an internal memory module to capture the {\bf emotion dynamics} during decoding. 
We simulate the process of expressing emotions as follows: there is an internal emotion state for each category before the decoding process starts; at each step the emotion state
decays by a certain amount; once the decoding process is completed, the emotion state should decay to zero indicating the emotion is completely expressed.

\begin{figure}[!htbp]
  \centering
  \includegraphics[width=1\linewidth]{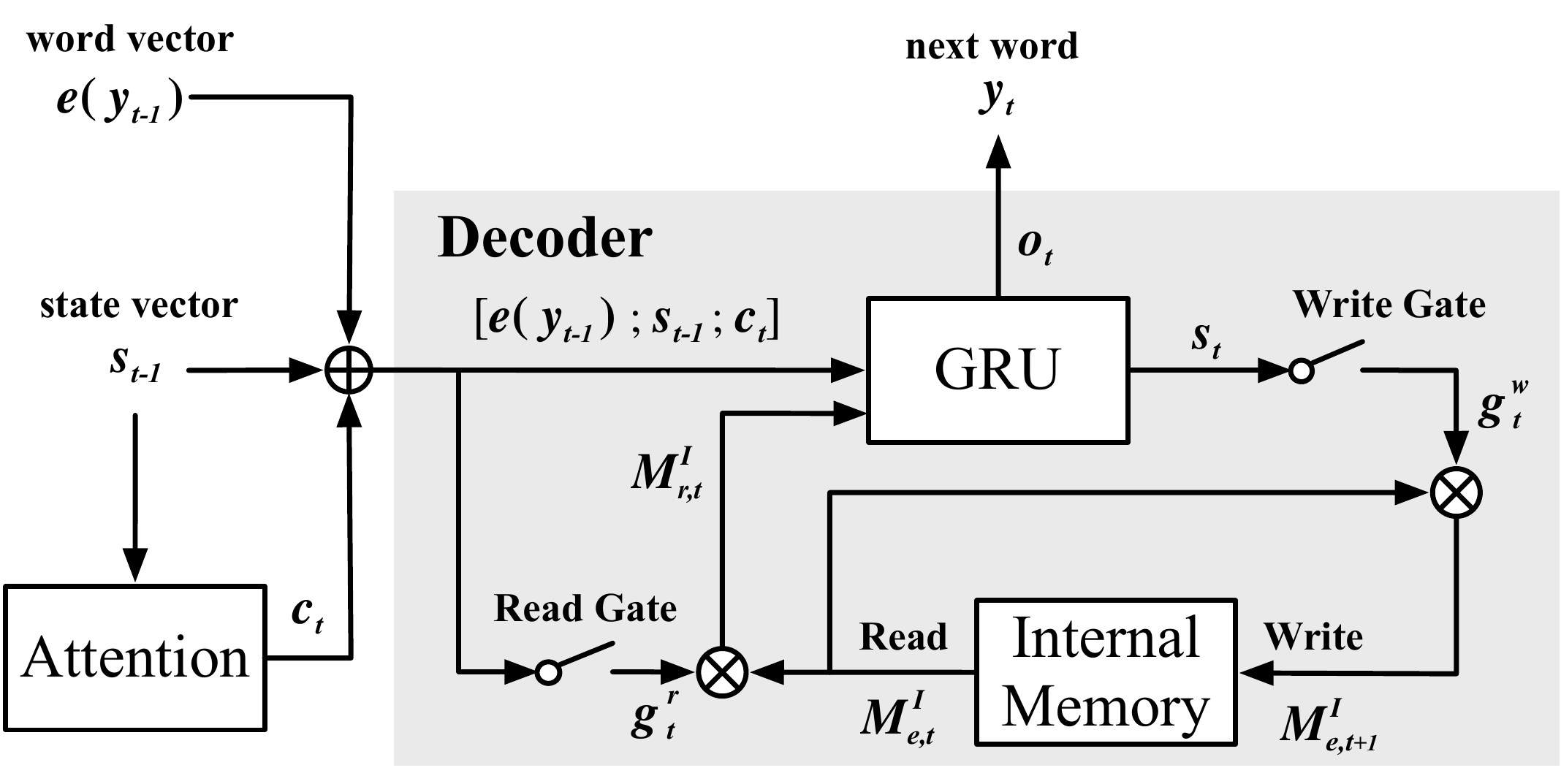}
  \caption{Data flow of the decoder with an internal memory. The internal memory $\bm{M}^I_{e,t}$ is read with the read gate $\bm{g}^r_t$ by an amount $\bm{M}^I_{r,t}$ to update the decoder's state, and the memory is updated to $\bm{M}^I_{e,t+1}$ with the write gate $\bm{g}^w_t$.
  }
  \label{fig:2}
\end{figure}

The detailed process of the internal memory module is illustrated in Figure \ref{fig:2}.
At each step $t$, ECM computes a read gate $\bm{g}^r_t$ with the input of the word embedding of the previously decoded word $\bm{e}(y_{t-1})$, the previous state of the decoder $\bm{s}_{t-1}$, and the current context vector $\bm{c}_t$. A write gate $\bm{g}^w_t$ is computed on the decoder's state vector $\bm{s}_t$. The read gate and write gate are defined as follows:
\begin{eqnarray}
\bm{g}^r_t&=&{\rm sigmoid}(\mathbf{W^r_g} [\bm{e}(y_{t-1}) ; \bm{s}_{t-1} ; \bm{c}_t]) ,\\
\bm{g}^w_t&=&{\rm sigmoid}(\mathbf{W^w_g} \bm{s}_t) .
\end{eqnarray}

The read and write gates are then used to read from and write into the internal memory, respectively. Hence, the emotion state is erased by a certain amount (by $\bm{g}^w_t$) at each step. At the last step, the internal emotion state will decay to zero. 
This process is formally described as below: 
\begin{eqnarray}
\bm{M}^I_{r,t}&=&\bm{g}^r_t \otimes \bm{M}^I_{e, t} ,\\
\bm{M}^I_{e,t+1}&=&\bm{g}^w_t \otimes \bm{M}^I_{e, t} ,
\label{eq:decay}
\end{eqnarray} 
where $\otimes$ is element-wise multiplication, $r/w$ denotes read/write respectively, and $I$ means Internal. 
GRU updates its state $\bm{s}_t$ conditioned on the previous target word $\bm{e}(y_{t-1})$, the previous state of the decoder $\bm{s}_{t-1}$, the context vector $\bm{c}_t$, and the emotion state update $\bm{M}^I_{r,t}$, as follows:
\begin{equation}
\bm{s}_t  =  \mathbf{GRU}(\bm{s}_{t-1},[\bm{c}_t ; \bm{e}(y_{t-1}) ; \bm{M}^I_{r,t}]) .
\end{equation}

Based on the state, the word generation distribution $\bm{o}_t$ can be obtained with Eq. \ref{eq:soft}, and the next word $y_t$ can be sampled.
After generating the next word, $\bm{M}^I_{e,t+1}$ is written back to the internal memory. Note that if Eq. \ref{eq:decay} is executed many times, it is equivalent to continuously multiplying the matrix, resulting in a decay effect since $0 \le {\rm sigmoid}(\cdot) \le 1$. This is similar to a DELETE operation in memory networks~\cite{miller2016key}.

\subsection{External Memory}

In the internal memory module, the correlation between the change of the internal emotion state and selection of a word is \textbf{implicit and not directly observable}.
As the emotion expressions are quite distinct with emotion words~\cite{xu2008} contained in a sentence, such as {\it lovely} and {\it awesome}, which carry strong emotions compared to generic (non-emotion) words, such as {\it person} and {\it day}, we propose an external memory module to model emotion expressions \textbf{explicitly} by assigning different generation probabilities to emotion words and generic words. Thus, the model can choose to generate words from an emotion vocabulary or a generic vocabulary.

\begin{figure}[htbp]
  \centering
  \includegraphics[width=1\linewidth]{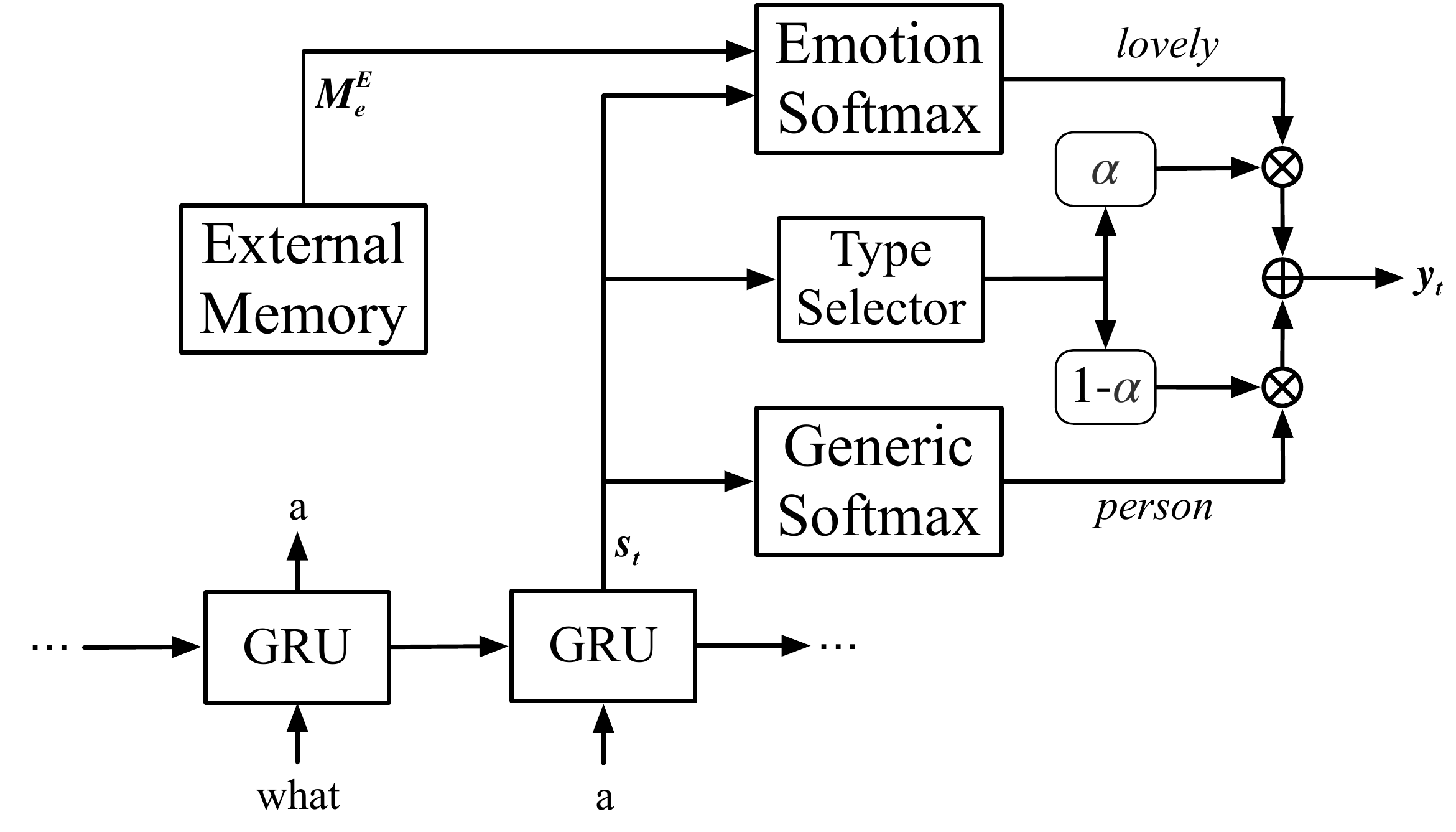}
  \caption{Data flow of the decoder with an external memory. The final decoding probability is weighted between the emotion softmax and the generic softmax, where the weight is computed by the type selector.
  }
  \label{fig:3}
\end{figure}

The decoder with an external memory is illustrated in Figure \ref{fig:3}. Given the current state of the decoder $\bm{s}_t$, the emotion softmax $P_e(y_t = w_e)$ and the generic softmax $P_g(y_t = w_g)$ are computed over the emotion vocabulary which is read from the external memory and generic vocabulary, respectively. The type selector $\alpha_t$ controls the weight of generating an emotion or a generic word.  Finally, the next word $y_t$ is sampled from the next word probability, the concatenation of the two weighted probabilities. The process can be formulated as follows:
\begin{eqnarray}
\alpha_t & = & {\rm sigmoid}(\mathbf{v_u}^\top \bm{s}_t) ,\\
P_g(y_t = w_g) & = & {\rm softmax}(\mathbf{W^o_g} \bm{s}_t) ,\\
P_e(y_t = w_e) & = & {\rm softmax}(\mathbf{W^o_e} \bm{s}_t) ,\\
\label{eq:decode_ot}
y_t \sim \bm{o}_t = P(y_t) & = & \small{\left[
\begin{aligned}
(1-\alpha_t)P_g(y_t = w_g) \\
\alpha_t P_e(y_t = w_e)
\end{aligned}
\right]
},
\end{eqnarray}
where $\alpha_t \in [0,1]$ is a scalar to balance the choice between an emotion word $w_e$ and a generic word $w_g$, 
$P_g/P_e$ is the distribution over generic/emotion words respectively,
and $P(y_t)$ is the final word decoding distribution. Note that the two vocabularies have no intersection, and 
the final distribution $P(y_t)$ is a concatenation of two distributions.

\subsection{Loss Function}
The loss function is the cross entropy error between the predicted token distribution $\bm{o}_t$ and the gold distribution $\bm{p}_t$ in the training corpus. Additionally, we apply two regularization terms: one on the internal memory, enforcing that the internal emotion state should decay to zero at the end of decoding, and the other on the external memory, constraining the selection of an emotional or generic word.  The loss on one sample $<\bm{X},\bm{Y}>$ ($\bm{X}=x_1,x_2,...,x_n$, $\bm{Y}=y_1,y_2,...,y_m$) is defined as: 
\begin{equation}
\label{eq:loss}
L(\theta)= -\sum_{t=1}^m \bm{p}_t {\rm log}(\bm{o}_t) - \sum_{t=1}^m q_t {\rm log}(\alpha_t) + \parallel \bm{M}_{e,m}^I \parallel ,
\end{equation}
where $\bm{M}_{e,m}^I$ is the internal emotion state at the last step $m$, $\alpha_t$ is the probability of choosing an emotion word or a generic word, and $q_t \in \{0,1\}$ is the true choice of an emotion word or a generic word in $\bm{Y}$. The second term is used to supervise the probability of selecting an emotion or generic word. And the third term is used to ensure that the internal emotion state has been expressed completely once the generation is completed.

\section{Data Preparation} 
Since there is no off-the-shelf data to train ECM, we firstly trained an emotion classifier using the NLPCC emotion classification dataset and then used the classifier to annotate the STC conversation dataset ~\cite{Shang2015Neural} to construct our own experiment dataset. 
There are two steps in the data preparation process:
\subsubsection{1. Building an Emotion Classifier.} 
We trained several classifiers on the NLPCC dataset and then chose the best classifier for automatic annotation. This dataset was used in challenging tasks of emotion classification in NLPCC2013\footnote{http://tcci.ccf.org.cn/conference/2013/} and NLPCC2014\footnote{http://tcci.ccf.org.cn/conference/2014/}, consisting of 23,105 sentences collected from Weibo. It was manually annotated with 8 emotion categories: {\it Angry}, {\it Disgust}, {\it Fear}, {\it Happy}, {\it Like}, {\it Sad}, {\it Surprise}, and {\it Other}. After removing the infrequent classes~({\it Fear} (1.5\%) and {\it Surprise} (4.4\%)), we have six emotion categories, i.e., {\it Angry}, {\it Disgust}, {\it Happy}, {\it Like}, {\it Sad} and {\it Other}.

We then partitioned the NLPCC dataset into training, validation, and test sets with the ratio of 8:1:1. Several emotion classifiers were trained on the filtered dataset, including a lexicon-based classifier~\cite{liu2012} (we used the emotion lexicon in~\cite{xu2008}), RNN~\cite{mikolov2010recurrent}, LSTM~\cite{hochreiter1997long}, and Bidirectional LSTM (Bi-LSTM)~\cite{graves2005bidirectional}. Results in Table \ref{tab:classifier} show that all neural classifiers outperform the lexicon-based classifier, and the Bi-LSTM classifier obtains the best accuracy of 0.623.

\begin{table} [!htp]
\centering
\begin{tabular}{l|c}
\hline
Method & Accuracy\\
\hline
Lexicon-based & 0.432 \\
RNN & 0.564 \\
LSTM & 0.594 \\
Bi-LSTM & 0.623 \\
\hline
\end{tabular}
\caption{Classification accuracy on the NLPCC dataset.}
\label{tab:classifier}
\end{table}

\subsubsection{2. Annotating STC with Emotion.}
\label{sec:classifier}
We applied the best classifier, Bi-LSTM, to annotate the STC Dataset with the six emotion categories. 
After annotation, we obtained an emotion-labeled dataset, which we call the {\it Emotional STC (ESTC) Dataset}. The statistics of the {\it ESTC Dataset} are shown in Table \ref{tab:dataset}. 
Although the emotion labels for {\it ESTC Dataset} are noisy due to automatic annotation, this dataset is good enough to train the models in practice. As future work, we will study how the classification errors influence response generation.

\begin{table} [!htp]
\centering
\begin{tabular}{|c|c|l|l|}
\hline
\multirow{7}{*}{{Training}} & {Posts} & \multicolumn{2}{c|}{217,905}\\
\cline{2-4}
& \multirow{6}{*}{{Responses}} &  Angry & 234,635\\
\cline{3-4}
& &  Disgust & 689,295\\
\cline{3-4}
& &  Happy & 306,364\\
\cline{3-4}
& &  Like & 1,226,954\\
\cline{3-4}
& &  Sad & 537,028\\
\cline{3-4}
& &  Other & 1,365,371\\
\hline
{Validation} & {Posts} & \multicolumn{2}{c|}{1,000} \\
\hline
{Test} & {Posts} & \multicolumn{2}{c|}{1,000} \\
\hline
\end{tabular}\caption{Statistics of the {\it ESTC Dataset}.}
\label{tab:dataset}
\end{table}

\section{Experiments} 
\subsection{Implementation Details}

We used Tensorflow\footnote{https://github.com/tensorflow/tensorflow}  to implement the proposed model\footnote{https://github.com/tuxchow/ecm}. The encoder and decoder have 2-layer GRU structures with 256 hidden cells for each layer and use different sets of parameters respectively. The word embedding size is set to 100. The vocabulary size is limited to 40,000. The embedding size of  emotion category is set to 100. The internal memory is a trainable matrix of size $6 \times 256$ and the external memory is a list of 40,000 words containing generic words and emotion words (but emotion words have different markers). To generate diverse responses, we adopted beam search in the decoding process of which the beam size is set to 20, and then reranked responses by the generation probability after removing those containing \emph{UNK}s, unknown words.

We used the stochastic gradient descent (SGD) algorithm with mini-batch. Batch size and learning rate are set to 128 and 0.5, respectively. To accelerate the training process, we trained a seq2seq model on the {\it STC} dataset with pre-trained word embeddings. And we then trained our model on the {\it ESTC} Dataset with parameters initialized by the parameters of the pre-trained seq2seq model. We ran 20 epoches, and the training stage of each model took about a week on a Titan X GPU machine.

\subsection{Baselines}
As aforementioned, this paper is the first work to address the emotion factor in large-scale conversation generation. 
We did not find closely-related baselines in the literature. Affect-LM \cite{GhoshCLMS17} cannot be our baseline because it is unable to generate responses of different emotions for the same post. Instead, it simply copies and uses the emotion of the input post. Moreover, it depends heavily on linguistic resources and needs manual parameter adjustments. 

Nevertheless, we chose two suitable baselines: a general seq2seq model~\cite{sutskever2014sequence}, and an emotion category embedding model (Emb) created by us where the emotion category is embedded into a vector, and the vector serves as an input to every decoding position, similar to the idea of user embedding in \cite{LiGBSGD16}. As emotion category is a high-level abstraction of emotion expressions, this is a proper baseline for our model.

\subsection{Automatic Evaluation}

\subsubsection{Metrics:}
As argued in \cite{LiuLSNCP16}, BLEU is not suitable for measuring conversation generation due to its low correlation with human judgment. We adopted perplexity to evaluate the model at the content level (whether the content is relevant and grammatical). To evaluate the model at the emotion level, we adopted emotion accuracy as the agreement between the expected emotion category (as input to the model) and the predicted emotion category of a generated response by the emotion classifier.

\begin{table} [!htp]
\centering
\begin{tabular}{l|l|l}
\hline
Method & Perplexity & Accuracy\\
\hline
Seq2Seq & 68.0 & 0.179 \\
Emb & 62.5 & 0.724 \\
\hline
ECM & 65.9 & {\bf 0.773} \\
w/o Emb & 66.1 & 0.753 \\
w/o IMem & 66.7 & 0.749 \\
w/o EMem & {\bf 61.8} & 0.731 \\

\hline
\end{tabular}
\caption{Objective evaluation with perplexity and accuracy.}
\label{tab:result}
\end{table}

\subsubsection{Results:}
The results are shown in Table \ref{tab:result}. As can be seen, ECM obtains the best performance in emotion accuracy, and the performance in perplexity is better than Seq2Seq but worse than Emb. This may be because the loss function of ECM is supervised not only on perplexity, but also on the selection of generic or emotion words (see Eq.\ref{eq:loss}). In practice, emotion accuracy is more important than perplexity considering that the generated sentences are already fluent and grammatical with the perplexity of 68.0.

In order to investigate the influence of different modules, we conducted ablation tests where one of the three modules was removed from ECM each time. As we can see, ECM without the external memory achieves the best performance in perplexity. 
Our model can generate responses without sacrificing grammaticality by introducing the internal memory, where the module can balance the weights between grammar and emotion dynamically. After removing the external memory, the emotion accuracy decreases the most, indicating the external memory leads to a higher emotion accuracy since it explicitly chooses the emotion words. Note that the emotion accuracy of Seq2Seq is extremely low because it generates the same response for different emotion categories.

\subsection{Manual Evaluation}

In order to better understand the quality of the generated responses from the content and emotion perspectives, we performed manual evaluation. Given a post and an emotion category, responses generated from all the models were randomized and presented to three human annotators.
\subsubsection{Metrics:}
Annotators were asked to score a response in terms of {\it Content} (rating  scale is 0,1,2) and {\it Emotion} (rating scale is 0,1), and also to state a preference between any two systems. {\it Content} is defined as whether the response is appropriate and natural to a post and could plausibly have been produced by a human, which is a widely accepted metric adopted by researchers and conversation challenging tasks, as proposed in \cite{Shang2015Neural}. {\it Emotion} is defined as whether the emotion expression of a response agrees with the given emotion category.

\subsubsection{Annotation Statistics:}
We randomly sampled 200 posts from the test set. For each model we generated 1,200 responses in total: for Seq2Seq, we generated the top 6 responses for each post, and for Emb and ECM, we generated the top responses corresponding to the 6 emotion categories.

We calculated the Fleiss' kappa \cite{fleiss1971measuring} to measure inter-rater consistency. Fleiss' kappa for {\it Content} and {\it Emotion} is 0.441 and 0.757, indicating ``Moderate agreement'' and ``Substantial agreement'' respectively.

\begin{table} [!htp]
\centering
\begin{tabular}{l|c|c|c|c|c|c}
\hline
Method (\%) & 2-1 & 1-1 & 0-1 & 2-0 & 1-0 & 0-0\\
\hline
Seq2Seq & 9.0 & 5.1 & 1.1 & 37.6 & 28.0 & 19.2\\
Emb & 22.8 & 9.3 & 4.3 & 27.1 & 19.1 & 17.4\\
ECM & {\bf 27.2} & {\bf 10.8} & 4.4 & 24.2 & 15.5 & 17.9\\
\hline
\end{tabular}
\caption{The percentage of responses in manual evaluation with the score of {\it Content}-{\it Emotion}. For instance, 2-1 means content score is 2 and emotion score is 1.}
\label{tab:percent}
\end{table}

\begin{figure*}[!htp]
  \centering
  \includegraphics[width=1.0\linewidth]{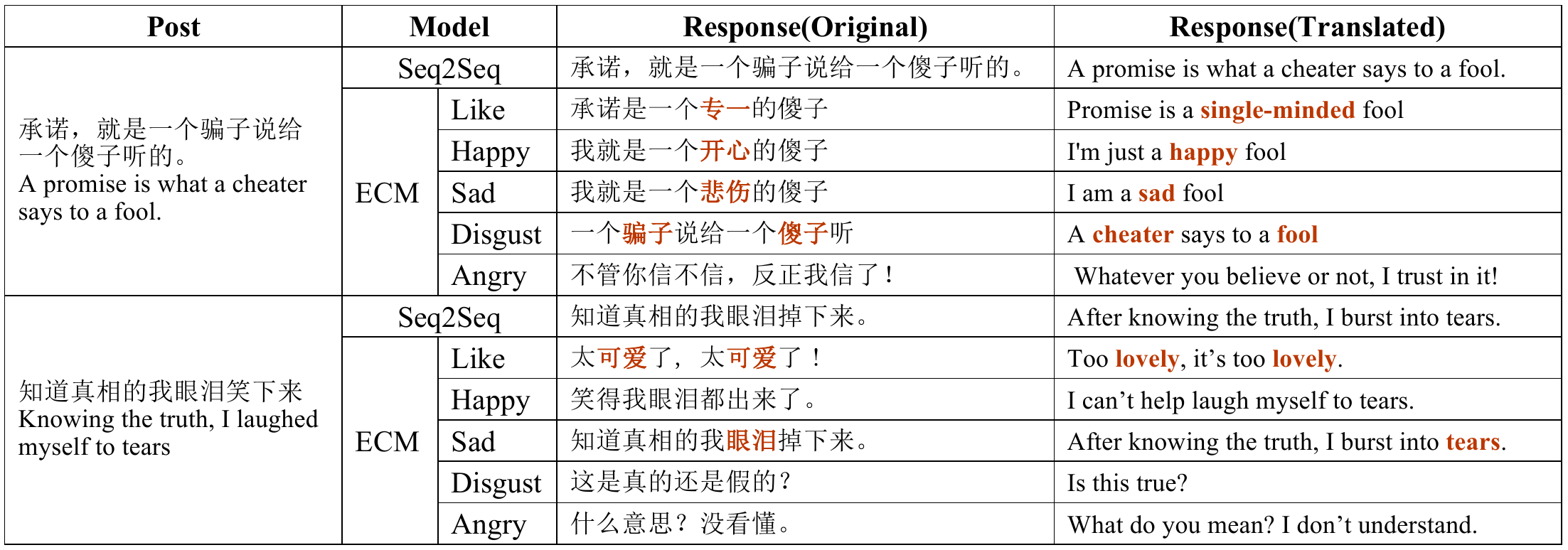}
  \caption{Sample responses generated by Seq2Seq and ECM (original Chinese and English translation, the colored words are the emotion words corresponding to the given emotion category). The corresponding posts did not appear in the training set.}
  \label{fig:sample}
\end{figure*}

\begin{table*} [!htp]
\centering
\begin{tabular}{l|l|l|l|l|l|l|l|l|l|l|l|l}
\hline
\multirow{2}{*}{Method} & \multicolumn{2}{c|}{Overall} & \multicolumn{2}{c|}{Like} & \multicolumn{2}{c|}{Sad} & \multicolumn{2}{c|}{Disgust} & \multicolumn{2}{c|}{Angry} & \multicolumn{2}{c}{Happy}  \\
\cline{2-13}
 & Cont. & Emot. & Cont. & Emot. & Cont. & Emot. & Cont. & Emot. & Cont. & Emot. & Cont. & Emot. \\
\hline
Seq2Seq & 1.255 & 0.152 & 1.308 & 0.337 & 1.270 & 0.077 & {\bf 1.285} & 0.038 & {\bf 1.223} & 0.052 & 1.223 & 0.257 \\
Emb & 1.256 & 0.363 & 1.348 & 0.663 & 1.337 & 0.228 & 1.272 & 0.157 & 1.035 & 0.162 & 1.418 & 0.607\\
ECM & {\bf 1.299} & {\bf 0.424} & {\bf 1.460} & {\bf 0.697} & {\bf 1.352} & {\bf 0.313} & 1.233 & {\bf 0.193} & 0.98 & {\bf 0.217} & {\bf 1.428} & {\bf 0.700} \\
\hline
\end{tabular}
\caption{Manual evaluation of the generated responses in terms of {\it Content} (Cont.) and {\it Emotion} (Emot.) 
. }
\label{tab:score}
\end{table*}
\subsubsection{Results:}
The results 
are shown in Table \ref{tab:score}.
ECM with all options outperforms the other methods in both metrics significantly~(2-tailed $t$-test, $p < 0.05$ for {\it Content}, and $p < 0.005$ for {\it Emotion}). After incorporating the internal memory and the external memory modules, the performance of ECM in {\it Emotion} is improved comparing to Emb, indicating our model can generate more explicit expressions of emotion. Besides, the performance in {\it Content} is improved from 1.256 of Emb to 1.299 of ECM, which shows the ability of ECM to control the weight of emotion and generate responses appropriate in content.

For all emotion categories, the performance of ECM in {\it Emotion} outperforms the other methods. However, the performances of ECM in {\it Content} is worse than baselines in {\it Disgust} and {\it Angry} categories, due to the fact that there are not sufficient training data for the two categories. For instance, the {\it Angry} category has 234,635 responses in our {\it ESTC Dataset}, much less than the other categories.

To evaluate whether ECM can generate responses that are appropriate not only in content but also in emotion, we present results in Table \ref{tab:percent} by considering content and emotion scores simultaneously\footnote{Note that {\it Content} and {\it Emotion} are two independent metrics.}. As we can see, 27.2\% of the responses generated by ECM have a {\it Content} score of 2 and an {\it Emotion} score of 1, while only 22.8\% for Emb and 9.0\% for Seq2Seq. These indicate that ECM is better in generating high-quality responses in both content and emotion.

\subsubsection{Preference Test:}

\begin{table} [!htp]
\centering
\begin{tabular}{l|cccc}
\hline
Pref. (\%) & Seq2Seq & Emb & ECM \\
\hline
Seq2Seq & - & 38.8 & 38.6 \\
Emb & 60.2 & - & 43.1 \\
ECM & {\bf 61.4} & {\bf 56.9} & - \\
\hline
\end{tabular}
\caption{Pairwise preference of the three systems.}
\label{tab:pref}
\end{table}

In addition, emotion models (Emb and ECM) are much more preferred than Seq2Seq, and ECM is also significantly (2-tailed t-test, $p < 0.001$) preferred by annotators against other methods as shown in Table \ref{tab:pref}. The diverse emotional responses are more attractive to users than the generic responses generated by the Seq2Seq model. And with the explicitly expressions of emotions as well as the appropriateness in content, ECM is much more preferred.

\subsection{Analysis of Emotion Interaction and Case Study}

\begin{figure}[htbp]
  \centering
  \includegraphics[width=0.8\linewidth]{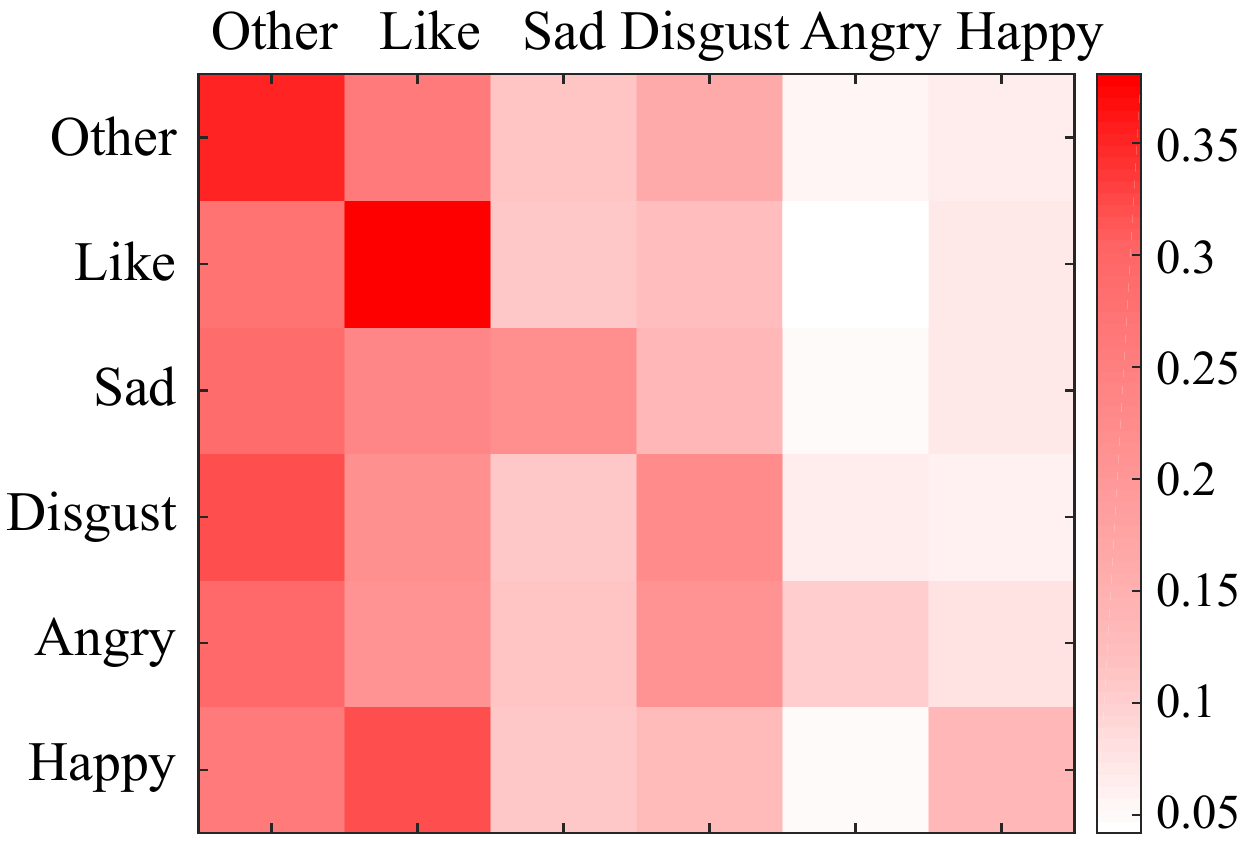}
  \caption{Visualization of emotion interaction. }
  \label{fig:EI}
\end{figure}

Figure \ref{fig:EI} visualizes the emotion interaction patterns of the posts and responses in the {\it ESTC Dataset}. An emotion interaction pattern (EIP) is defined as $<e_p,e_r>$, the pair of emotion categories of the post and its response. The value of an EIP is the conditional probability $P(e_r|e_p) = P(e_r,e_p) / P(e_p)$. 
An EIP marked with a darker color occurs more frequently than a lighter color. From the figure, we can make a few observations. \textbf{First}, frequent EIPs show that there are some major responding emotions given a post emotion category. For instance, when a post expresses {\it Happy}, the responding emotion is typically {\it Like} or {\it Happy}.
\textbf{Second}, the diagonal patterns indicate emotional empathy, a common type of emotion interaction. 
\textbf{Third}, there are also other EIPs for a post, indicating that emotion interactions in conversation are quite diverse, as mentioned earlier.  
Note that class {\it Other} has much more data than other classes (see Table \ref{tab:dataset}), indicating that EIPs are biased toward this class (the first column of Figure \ref{fig:EI}), due to the data bias and the emotion classification errors. 

We present some examples in Figure \ref{fig:sample}. As can be seen, for a given post, there are multiple emotion categories that are suitable for its response in conversation. Seq2Seq generates a response with a random emotion. ECM can generate emotional responses conditioned on every emotion category. All these responses are appropriate to the post, indicating the existence of multiple EIPs and the reason why an emotion category should be specified as an input to our system.  

We can see that ECM can generate appropriate responses if the pre-specified emotion category and the emotion of the post belong to one of the frequent EIPs. Colored words show that ECM can explicitly express emotion by applying the external memory which can choose a generic (non-emotion) or emotion word during decoding.
For low-frequency EIPs such as $<{\it Happy},{\it Disgust}>$ and $<{\it Happy},{\it Angry}>$ as shown in the last two lines of Figure \ref{fig:sample}, responses are not appropriate to the emotion category due to the lack of training data and/or the errors caused by the emotion classifier.

\section{Conclusion and Future Work} 
\label{sec:conclusion}
In this paper, we proposed the Emotional Chatting Machine (ECM) to model the emotion influence in large-scale conversation generation. Three mechanisms were proposed to model the emotion factor, including emotion category embedding, internal emotion memory, and external memory. Objective and manual evaluation show that ECM can generate responses appropriate not only in content but also in emotion.

In our future work, we will explore emotion interactions with ECM: instead of specifying an emotion class, the model should decide the most appropriate emotion category for the response. However, this may be challenging since such a task depends on the topics, contexts, or the mood of the user. 

\section{Acknowledgments}
This work was partly supported by the National Science Foundation of China under grant No.61272227/61332007, and a joint project with Sogou. We would like to thank our collaborators, Jingfang Xu and Haizhou Zhao.

\bibliography{Zhou-Huang}
\bibliographystyle{aaai}

\end{document}